\pdfoutput=1

\documentclass[11pt]{article}

\usepackage{ACL2023}

\usepackage{times}
\usepackage{latexsym}

\usepackage{booktabs}
\usepackage{tabularx}
\usepackage{enumitem}
\usepackage[dvipsnames]{xcolor}
\usepackage{graphicx}
\usepackage{amssymb}
\usepackage{adjustbox}

\usepackage[T1]{fontenc}

\usepackage[utf8]{inputenc}

\usepackage{microtype}

\usepackage{inconsolata}

\usepackage{placeins}

\title{LR-Sum: Summarization for Less-Resourced Languages}

\author{Chester Palen-Michel \and Constantine Lignos \\
  Mitchom School of Computer Science \\
  Brandeis University \\
  \texttt{\{cpalenmichel, lignos\}@brandeis.edu} \\
  }

\begin{document}
\maketitle
\begin{abstract}
We introduce LR-Sum, a new permissively-licensed dataset created with the goal of enabling further research in automatic summarization for less-resourced languages.
LR-Sum contains human-written summaries for 40 languages, many of which are less-resourced. 
We describe our process for extracting and filtering the dataset from the Multilingual Open Text corpus \citep{palen-michel-etal-2022-multilingual}.
The source data is public domain newswire collected from from Voice of America websites, and LR-Sum is released under a Creative Commons license (CC BY 4.0), making it one of the most openly-licensed multilingual summarization datasets.
We describe abstractive and extractive summarization experiments to establish baselines and discuss the limitations of this dataset.
\end{abstract}

\section{Introduction}
Datasets for automatic summarization have historically focused largely on English, and while there has recently  been a greater focus on datasets that include other languages \citep{cao2020multisumm, giannakopoulos2015multiling, giannakopoulos2017multiling, hasan-etal-2021-xl, scialom-etal-2020-mlsum}, there still remains a need for high-quality summarization data for less-resourced languages.
Datasets with human-written summaries are important for both training statistical summarization models and for automatic evaluation of them.
While recently there have been a growing number of multilingual summarization datasets, many are relatively small, have limited language coverage, have restrictive licenses, or a combination of these drawbacks.

\begin{table}[tb]
\centering
\small
\begin{tabularx}{\columnwidth}{X}
\toprule
Summary: \\
\textcolor{ForestGreen}{Fuad Huseyîn} \textcolor{Mahogany}{li civîna} \textcolor{RoyalBlue}{NY} \textcolor{BurntOrange}{jiber êrîşê Enqere tawanbar kir}; \textcolor{RoyalPurple}{daxwaza tazmînat û lêpirsîneke navneteweyî kir}. \\
\midrule
Article: \\
Wezîrê derve yê Îraqê \textcolor{ForestGreen}{Fuad Huseyîn} doh Sêşemê \textcolor{Mahogany}{li civîna} awarte ya Civata Ewlekarîyê ya \textcolor{RoyalBlue}{Neteweyên Yekbûyî (NY)}, daxwaza vekişîna hêzên Tirkîyê ji axa Îraqê kir. ``Em hebûna neqanûnî ya hêzên artêşa Tirkîyê li ser axa Îraqê şermezar dikin,'' \textcolor{ForestGreen}{Huseyîn} got. 
Civata Ewlekarîyê ya \textcolor{RoyalBlue}{Neteweyên Yekbûyî (NY)} ser daxwaza Îraqê kom bû, di derbarê êrîşa hefteya borî ya li Duhokê hat kirin û di encamê de 9 kes hatibûn kuştin û 23 kesên din jî birîndar bûbûn. Îraq \textcolor{BurntOrange}{jiber êrîşa kujer hêzên Tirkiyê tawanbar} dike û wezîrê derve \textcolor{ForestGreen}{Huseyîn} çû New Yorkê da ku beşdarî \textcolor{Mahogany}{civîna} awarte ya \textcolor{RoyalBlue}{NY} bibe. \textcolor{ForestGreen}{Fuad Huseyîn} \textcolor{Mahogany}{li civînê} ser navê Bexdayê, \textcolor{BurntOrange}{jiber êrîşê Enqere tawanbar kir} û \textcolor{RoyalPurple}{daxwaza tazmînatê û lêpirsîneke navneteweyî kir}.
[...] \\
\bottomrule
\end{tabularx}
\caption{Example summary and article pair from Kurmanji Kurdish. Colors mark approximate content equivalence between summary and a portion of the article.}
\label{tab:kurdish-example}
\end{table}

\begin{table}[tb]
\centering
\small
\begin{tabularx}{\columnwidth}{X}
\toprule
Summary: \\
\textcolor{Mahogany}{First-ever aerial census} will be \textcolor{Magenta}{conducted simultaneously} \textcolor{BurntOrange}{across five states} to determine \textcolor{ForestGreen}{elephant} \textcolor{RoyalPurple}{migration patterns} and \textcolor{RoyalBlue}{numbers} \\
\midrule
Article: \\
\textcolor{BurntOrange}{Five southern African countries}, with more than half the continent's \textcolor{ForestGreen}{elephants}, are conducting a \textcolor{Mahogany}{first-ever aerial census} to determine the \textcolor{ForestGreen}{elephant} \textcolor{RoyalBlue}{population} and how to protect it.
Light aircraft will \textcolor{Magenta}{fly simultaneously} across the plains of \textcolor{BurntOrange}{Angola, Botswana, Namibia, Zambia and Zimbabwe} — in a conservation area known as the Kavango-Zambezi Trans-frontier Conservation Area (KAZA) — in an exercise that will run until October 20.
[...]
We hope to see what the results come up with,'' Ives said. ``What we will be interested in seeing is not only \textcolor{RoyalBlue}{how many elephants there are} but the distribution, therefore, and what the likelihood of those \textcolor{ForestGreen}{elephants} \textcolor{RoyalPurple}{moving between countries} is.
\\
\bottomrule
\end{tabularx}
\caption{Example summary and article pair from English. Colors mark approximate content equivalence between summary and a portion of the article.}
\label{tab:english-example}
\end{table}

In this paper, we present LR-Sum, a new 40-language summarization dataset with a focus on less-resourced languages.\footnote{There is no definitive definition for less-resourced \citep{liu-etal-2022-always} and we take the view that less-resourced can depend on the intersection of many factors \cite{lignos-etal-2022-toward}, including what task a dataset is created for.}
We created it with the goal of providing high-quality, human-written summaries in as many languages as possible.
The collection of curated and filtered summaries that comprise LR-Sum are licensed using a Creative Commons Attribution license (CC BY 4.0), and the articles that it was collected from are in the public domain.
This allows LR-Sum to be distributed freely and annotated without restriction, unlike many summarization datasets which use copyrighted material, often redistributed without appropriate licensing.
For many of the languages in LR-Sum, this is the largest collection of summarization data with such a permissive license.

Tables~\ref{tab:kurdish-example} and~\ref{tab:english-example} show example article-summary pairs from LR-Sum and highlight how similar content in the summary is not merely simple extraction from the text.
Results of experiments described in Section \ref{sec:experiments} show that for many less-resourced languages, the task of producing summaries remains challenging, enabling LR-Sum to serve as a benchmark of progress. 
LR-Sum is released via GitHub at \url{https://github.com/bltlab/lr-sum}.


\section{Related Work}

In this section, we briefly list existing English and multilingual summarization datasets and discuss work in dataset creation for less-resourced languages more generally.

\subsection{English Summarization Datasets}

\textbf{Document Understanding Conference (DUC)}\footnote{\url{http://duc.nist.gov/}} \citep{harman-over-2004-effects,dang-2006-duc} create English summarization datasets for evaluations. 

\textbf{The NYT Annotated Corpus} \citep{sandhaus2008new} is a corpus of New York Times articles and 600k summaries written by library scientists. 

\textbf{CNN/Daily Mail} \citep{hermann2015teaching} was originally created for question answering, but \citet{nallapati-etal-2016-abstractive-fix} adapt this dataset for summarization. 

\textbf{XSum} \citep{narayan-etal-2018-dont} uses the first sentence ``story body introduction'' tag of a BBC article as the summary and the remainder of the text as the article and show that XSum favors abstractive summaries.

\subsection{Multilingual Summarization Datasets}
\textbf{MLSUM} \citep{scialom-etal-2020-mlsum} is an extension of the CNN/Daily Mail dataset for five languages: French, German, Spanish, Turkish, and Russian. 

\textbf{MultiLing} \citep{giannakopoulos2015multiling, giannakopoulos2017multiling} is a shared task that focuses on multilingual summarization covering upwards of 40 languages, but the dataset size is somewhat limited, with training sets of only around 10,000 articles in total.

\textbf{XL-Sum} \citep{hasan-etal-2021-xl} includes 44 languages, many of which are less-resourced languages, by scraping BBC News and making use of bullet points as summaries. 
XL-Sum has a more restrictive license than LR-Sum. 

\textbf{MassiveSumm} \citep{varab-schluter-2021-massivesumm} is a very large web-scraped summarization corpus that covers the majority of languages covered both in our dataset, LR-Sum, and also XL-Sum, and it does so in larger quantities. 
However, MassiveSumm cannot be easily redistributed due to copyright and being scraped from various news sites. 
MassiveSumm's GitHub README contains the disclaimer ``The data is noisy and recall-oriented.''\footnote{\url{https://github.com/danielvarab/massive-summ}}

\textbf{MultiSumm} \citep{cao2020multisumm} creates summaries from titles for Bosnian and Croatian.

\subsection{Data for Less-Resourced Languages}

A number of other text corpora have been created for less-resourced languages for summarization and other tasks.
\citet{abdulrahman2019developing} create a Kurdish corpus of textbooks.
\citet{vasili2018study} conduct a study on summarization of Albanian and build a small dataset.
\citet{malajyan-2020-arpa} create a corpus of paraphrases in Armenian. 
\citet{niyongabo-etal-2020-kinnews} create a corpus for classification of news in Kinyarwanda and Kirundi. 
\citet{Azime-2021-amhnews} create a news dataset in Amharic.
\citet{marivate-etal-2020-investigating} investigate corpus creation for Setswana and Sepedi. 
\citet{koto-etal-2020-liputan6} create a summarization corpus for Indonesian with over 200k articles, but it has significant license restrictions.
\citet{das-bandyopadhyay-2010-topic} create a system for opinion article summarization for Bangla.
\citet{nguyen-etal-2020-study} create a dataset and experiment with sentence compression in Vietnamese.
\citet{Amiha2017AutomaticTS} create an extractive summarization system and evaluate on a small dataset in Tigrinya.
\citet{jaruskulchai-kruengkrai-2003-practical} create an extractive model for Thai, and \citet{chumpolsathien_2020} create a large-scale dataset for Thai summarization.  
\citet{buoy-2021-khmer} explore text classification with Khmer.

\textbf{Multilingual Open Text (MOT)} \citep{palen-michel-etal-2022-multilingual} is a corpus collected from the websites of Voice of America, an international news service funded by the U.S. Government providing news articles and other short snippets like audio and image descriptions.  
Our work creates a summarization dataset for the majority of the languages within MOT. 
MOT has a permissive license (CC BY 4.0), and the original source articles are in the public domain. 
By comparison, many of the multilingual datasets derived from privately funded news sources like CNN or BBC News were collected from copyrighted data without the copyright owner's permission, limiting legal distribution.
XL-Sum's license is CC BY-NC-SA 4.0, which restricts commercial usage.
MOT contains news text data for many less-resourced languages, some of which overlap with XL-Sum and some of which are complementary. 
We discuss which languages are present in LR-Sum vs XL-Sum in more detail in Section~\ref{section:dataset-description}.

\section{LR-Sum: Dataset}

\subsection{Methodology}
The approach for creating LR-Sum is to leverage the coverage of less-resourced languages in MOT to construct a summarization dataset.
MOT \citep{palen-michel-etal-2022-multilingual} semi-regularly releases new versions of the dataset as new articles are published on Voice of America's website. 
We use MOT release v1.6 from October 1, 2022 for the creation of LR-Sum. 
Only the content type of ``article'' is included in LR-Sum since the categories of photo, audio, etc. already tend to be short snippets describing content, which typically are too short to make useful article-summary pairs.

While bold text or bullet points are used in some other summarization datasets \citep{hasan-etal-2021-xl, hermann2015teaching, narayan-etal-2018-dont}, these ways of extracting summaries are not available in VOA articles.
Instead a description field is present for VOA articles.
This description field in VOA new articles can be noisy. 
While it is generally used to give a brief summary of the article contents, there are numerous instances where the description contains the first few lines of the article, information about the authors, or general information about what VOA is.

A number of filtering steps are taken to ensure high-quality summaries.
First, we filter to ensure that the description field has content and that the content of the description field is at least 10 tokens long.\footnote{All tokenization comes from the tokenizers used in the creation of the MOT corpus.}
Then, we filter out any articles that do not have a minimum of 10 sentences.
We also filter by total number of tokens to remove outlier articles with fewer than 30 or more than 6,000 tokens.

When an article does not have a human-written summary, the description field simply contains the first few sentences.
Because ellipses can signal that the description is just a copy of the first few sentences of the article, we also filter out all descriptions that end with ellipses. 
We further remove these instances from the dataset by limiting token overlap of the description and the first 3 sentences to 85\%.\footnote{This cutoff was chosen based on manual review of summaries in multiple languages.}
With the goal of keeping LR-Sum from being purely extractive, we also block descriptions where an oracle extractive approach selecting the best sentence in the article produces a ROUGE-2 score above 95.

We manually created a list of 254 sentences to remove from summaries based on strings that appear the most frequently in the description field.
Examples include ``Amerikan basınında haftaiçi hergün öne çıkan başlıkları Amerika'nın Sesi'nde bulubilirsiniz'' (``You can find the highlights of the American press every weekday on Voice of America'' in Turkish)  or ``Këtë javë në Uashington'' (``Live from Washington'' in Albanian).\footnote{The list of excluded sentences is released with the dataset.}

While MOT includes data in the Lingala and Oromo languages, we do not include them in LR-Sum since fewer than 100 articles made it through our filtering process. 
Lingala had only 3 articles, and Oromo 29.
MOT also includes data in Bambara, but it contains so few articles that none made it through the filtering process.

\subsection{Dataset Description}
\label{section:dataset-description}
\begin{figure}[tb]
\centering
\includegraphics[width=\linewidth]{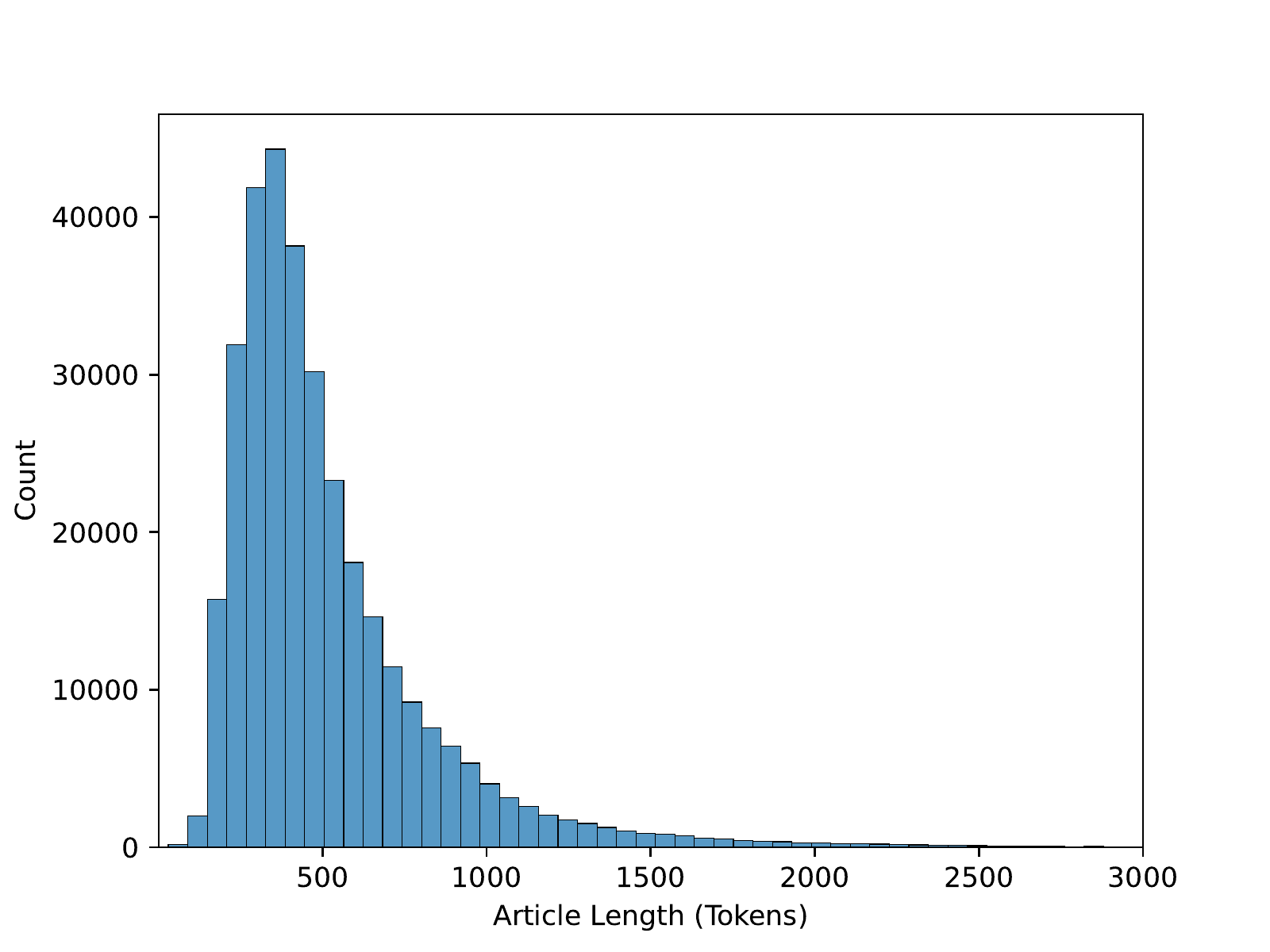}
\caption{Histogram of articles lengths in tokens.}
\label{fig:articles-histo}
\end{figure}

\begin{figure}[tb]
\centering
\includegraphics[width=\linewidth]{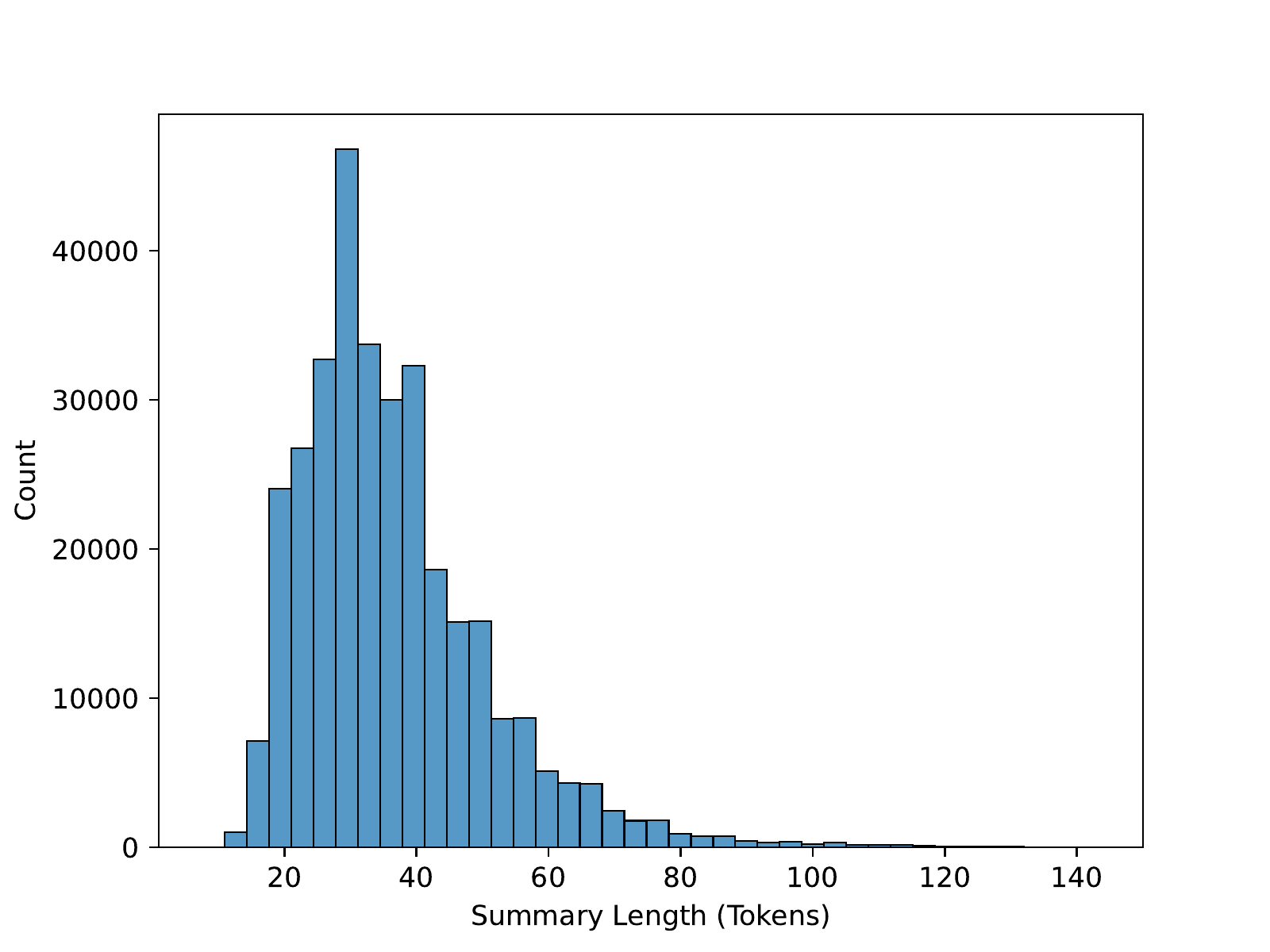}
\caption{Histogram of summary lengths in tokens.}
\label{fig:summaries-histo}
\end{figure}

LR-Sum includes 40 languages in total.
We show various statistics of the dataset in Table \ref{table:stats}. 
Figure~\ref{fig:articles-histo} provides a histogram of article lengths, and Figure~\ref{fig:summaries-histo} provides a histogram of summary lengths.

We measure mean length of articles and summaries in token counts. 
Compression is 1 - the ratio between summary length and article length as used by \citet{bommasani-cardie-2020-intrinsic} and \citet{hasan-etal-2021-xl}.
Mean novelty is the proportion of tokens in the summary that do not occur in the article. 
LR-Sum's measures are comparable with MLSUM \citep{scialom-etal-2020-mlsum} and XL-Sum \citep{hasan-etal-2021-xl} for languages shared between datasets.
The overall mean article length for LR-Sum is 520.7 and the overall mean summary length is 36.5.
For comparison, MLSUM's English section has a mean article length of 709.2 and mean summary length of 55.6, while their Turkish section has mean article length of 309.1 and mean summary length of 22.8. 

LR-Sum includes fourteen languages that are not covered by XL-Sum.
However, Dari Persian and Kinyarwanda are quite close to Persian Farsi and Kirundi, which are contained in XL-Sum.
Seven of the remaining twelve languages have more than 1,000 article-summary pairs for training: Albanian, Bosnian, Khmer, Sorani Kurdish, Lao, Macedonian, and Northern Ndebele. 
Armenian, Georgian, Haitian Creole, and Shona have fewer than 1,000 training examples. 
Tibetan and Greek have fewer than 1,000 article-summary pairs overall, which is not enough for training and test splits.  Instead, the Tibetan and Greek data could still be useful as a test set for automatic evaluation of models built for those languages or used in few-shot training.  

LR-Sum includes languages which can be complementary to existing resources. 
For example, LR-Sum includes almost twice as many articles in Burmese as XL-Sum. 
For many languages (i.e. Turkish, Azerbaijani, Persian, Korean) adding LR-Sum to XL-Sum results in more than double the amount of data available in XL-Sum alone.

LR-Sum also has some unique subdivisions and special focuses for certain languages. 
Its English section can be subdivided into Zimbabwe and Cambodia-focused sections. 
Similarly, the French and Portuguese found in LR-Sum tends to be news focused on Africa.
Chinese is divided into simplified and traditional varieties. 
Kurdish is subdivided into the Kurmanji and Sorani dialects. 
LR-Sum separates Farsi and Dari as separate languages based on their provenance from separate VOA sites, despite their being largely mutually intelligible.

\begin{table*}[tb!]
\centering
\begin{tabular}{llrrrrrr}
\toprule
            & ISO 639-3         & Mean    & Mean    &              &         &         &        \\
            & Language          & Article & Summary &              & Mean    & Vocab.  & Article\\
Language    & Code & Length     & Length  & Compression  & Novelty & Size    & Count  \\
\midrule
Albanian & sqi & 503.30 & 21.23 & .9578 & .2349 & 204,334 & 22,890 \\
Amharic & amh & 291.47 & 25.52 & .9124 & .4781 & 16,833 & 154 \\
Armenian & hye & 321.39 & 24.43 & .9240 & .3582 & 53,659 & 1,920 \\
Azerbaijani & aze & 390.86 & 14.98 & .9617 & .2915 & 178,330 & 8,108 \\
Bangla & ben & 310.45 & 29.23 & .9058 & .1032 & 27,288 & 715 \\
Bosnian & bos & 493.40 & 20.29 & .9589 & .2367 & 288,205 & 14,559 \\
Burmese & mya & 973.14 & 35.19 & .9638 & .1906 & 598,594 & 9,901 \\
Dari Persian & prs & 426.93 & 27.17 & .9364 & .2442 & 101,723 & 15,046 \\
English & eng & 717.98 & 32.11 & .9553 & .2053 & 194,901 & 38,697 \\
French & fra & 430.05 & 24.77 & .9424 & .1101 & 41,642 & 2,126 \\
Georgian & kat & 419.85 & 14.84 & .9647 & .2265 & 73,081 & 1,511 \\
Greek & ell & 482.42 & 12.96 & .9731 & .2442 & 28,976 & 583 \\
Haitian Creole & hat & 445.92 & 26.49 & .9406 & .1943 & 27,128 & 1,452 \\
Hausa & hau & 375.16 & 24.91 & .9336 & .2196 & 11,718 & 390 \\
Indonesian & ind & 363.69 & 20.06 & .9448 & .2069 & 39,907 & 1,968 \\
Khmer & khm & 896.77 & 32.76 & .9635 & .0764 & 54,986 & 4,860 \\
Kinyarwanda & kin & 351.41 & 18.34 & .9478 & .4274 & 39,678 & 698 \\
Korean & kor & 437.81 & 30.81 & .9296 & .4189 & 425,980 & 13,123 \\
Kurdish & kur & 541.53 & 22.36 & .9587 & .2781 & 128,429 & 4,021 \\
Lao & lao & 378.93 & 24.99 & .9340 & .1162 & 86,992 & 14,955 \\
Macedonian & mkd & 407.68 & 19.91 & .9512 & .3074 & 66,815 & 2,223 \\
Mandarin Chinese & cmn & 781.43 & 53.64 & .9314 & .2472 & 143,505 & 4,586 \\
Northern Ndebele & nde & 304.58 & 20.27 & .9335 & .2889 & 122,312 & 2,739 \\
Pashto & pus & 459.74 & 33.58 & .9270 & .2111 & 152,499 & 21,067 \\
Persian Farsi & fas & 512.21 & 31.08 & .9393 & .0870 & 126,339 & 13,429 \\
Portuguese & por & 489.19 & 18.23 & .9627 & .1637 & 46,578 & 1,643 \\
Russian & rus & 622.29 & 14.59 & .9766 & .2958 & 273,560 & 13,514 \\
Serbian & srp & 348.44 & 20.24 & .9419 & .4427 & 145,175 & 6,217 \\
Shona & sna & 276.12 & 17.47 & .9367 & .3189 & 45,808 & 1,383 \\
Somali & som & 463.87 & 24.73 & .9467 & .2599 & 12,736 & 165 \\
Spanish & spa & 651.14 & 32.13 & .9507 & .2116 & 66,094 & 3,544 \\
Swahili & swh & 361.96 & 24.53 & .9322 & .1777 & 23,110 & 588 \\
Thai & tha & 406.96 & 25.11 & .9383 & .3472 & 35,823 & 3,278 \\
Tibetan & bod & 904.99 & 62.44 & .9310 & .0357 & 6,886 & 182 \\
Tigrinya & tir & 281.30 & 13.02 & .9537 & .3217 & 10,156 & 115 \\
Turkish & tur & 447.94 & 23.67 & .9472 & .2915 & 308,870 & 35,839 \\
Ukrainian & ukr & 487.99 & 18.20 & .9627 & .2545 & 163,270 & 7,229 \\
Urdu & urd & 651.21 & 37.65 & .9422 & .0609 & 108,357 & 13,558 \\
Uzbek & uzb & 425.33 & 20.58 & .9516 & .3130 & 211,099 & 11,959 \\
Vietnamese & vie & 670.42 & 25.37 & .9622 & .1149 & 198,478 & 14,595 \\ 
\midrule
& & & & \multicolumn{3}{r}{Total Article Count} & 315,530 \\
\bottomrule
\end{tabular}
\caption{Metrics across languages in the LR-Sum Dataset.
Compression ratio is the ratio of article length to summary length.
Mean novelty is the mean proportion of tokens in the summary that do not occur in the article.
Vocabulary is the number of unique tokens (types).
All measures are computed using tokens.}
\label{table:stats}
\vspace{12pt}
\end{table*}

\subsection{Dataset Splits}
\label{sec:splits}
We report the size of the dataset splits for LR-Sum in Appendix \ref{tab:splits}. 
Splits are 80\% train, 10\% validation, and 10\% test, except for languages where the number of examples was quite small. 
To ensure enough test and validation data when possible,  in cases where the total was below 4,000 examples, we took 500 for validation and test each and left the rest for training. 
For languages where the total number of examples was fewer than 1,000, we only created test sets and did not create training or validation data (Amharic, Bangla, Greek, Hausa, Kinyarwanda, Somali, Swahili, Tibetan, and Tigrinya).

\section{Experiments}
\label{sec:experiments}
\subsection{Methodology}
We conduct three experiments to demonstrate the usefulness of LR-Sum and establish baseline performance on the dataset. 
For all abstractive models trained, we use mT5 \citep{xue-etal-2021-mt5} as the base model.
We report ROUGE-1 and 2 (R1, R2) and ROUGE-L \citep[RL;][]{lin-2004-rouge} scores.\footnote{We use \citet{hasan-etal-2021-xl}'s adaptation, but for consistency across languages, we do not make use of any of the stemmers available in the reported scores.}\textsuperscript{,}\footnote{All reported results are from single runs.}

\noindent
1. We train individual baseline models for 12 less-resourced languages that are unique to LR-Sum and not present in XL-Sum.\footnote{We treat Kurmanji and Sorani varieties of Kurdish separately here since they use different scripts and refer to them as kur-k for Kurmanji and kur-s for Sorani in results tables.}

\noindent
2. We conduct a series of experiments with extractive models for the less-resourced languages unique to LR-Sum.

\noindent
3. We train a multilingual model using the concatenation of LR-Sum and XL-Sum training sets and compare with using a multilingual model checkpoint trained on XL-Sum alone.
For this experiment, we evaluate both models on LR-Sum's test sets and evaluate on all less-resourced languages.

\subsubsection{Individual Models}
We fine-tune 12 models for each of the less-resourced languages not present in XL-Sum. 
We use mT5 \cite{xue-etal-2021-mt5} as the base model.
For these experiments, we use the same training script as \citet{hasan-etal-2021-xl}, which is a modified version of a script from the Hugging Face Transformers Library \citep{wolf-etal-2020-transformers}.
We use the same hyper-parameter settings as \citet{hasan-etal-2021-xl}.
The details of hyper-parameters can be found in Appendix \ref{sec:hyperparams-mono}.

\subsubsection{Extractive Baselines}
We conducted experiments to determine whether extractive approaches might work better given the small training set sizes.
Previous work by \citet{nallapati2017summarunner,narayan-etal-2018-ranking,zhang-etal-2018-neural} and \citet{scialom-etal-2020-mlsum}, among others,  has shown lead-3 (the first three sentences of the article) to be a strong summarization baseline. 
To demonstrate the strongest possible extractive performance, we also report the \texttt{oracle}, which here is simply selecting the single sentence in the article which produces the highest ROUGE score.
We additionally report results for LexRank \citep{erkan2004lexrank} and Luhn \cite{luhn1958automatic} extractive methods.
For implementations of these extractive approaches, we used sumy\footnote{\url{https://miso-belica.github.io/sumy/}} \citep{belica2013sumy}.
The sentence segmentation and tokenizations from the MOT corpus were used for the extractive approaches requiring segmentation and tokenization. 

\subsubsection{Multilingual Models}
Following \citet{hasan-etal-2021-xl}'s reported better performance with multilingual training, we train a multilingual model but instead with the concatenation of the training sets of LR-Sum and XL-Sum.
In this experiment we also use the same modified Hugging Face script \citep{wolf-etal-2020-transformers} that \citet{hasan-etal-2021-xl} use for training along with the same hyper-parameters as \citet{hasan-etal-2021-xl} used for multilingual training.
Hyper-parameter settings can be found in Appendix \ref{sec:hyperparams-multi}.

\section{Results and Discussion}
\label{sec:results}

\begin{table*}[tb!]
\resizebox{\linewidth}{!}{
\begin{tabular}{@{}lrrrrrrrrrrrrrrrr@{}}
\toprule
 & \multicolumn{3}{c}{Oracle} & \multicolumn{3}{c}{Lead3} & \multicolumn{3}{c}{LexRank} & \multicolumn{3}{c}{Individual Models} & \multicolumn{3}{c}{Multilingual Model} \\ \cmidrule(lr){2-4} \cmidrule(lr){5-7} \cmidrule(lr){8-10} \cmidrule(lr){11-13} \cmidrule(lr){14-16}
Lang. & R1 & R2 & RL & R1 & R2 & RL & R1 & R2 & RL & R1 & R2 & RL & R1 & R2 & RL \\ \midrule
sqi & 43.9 & 29.8 & 39.6 & 19.5 & 6.7 & 15.0 & 19.6 & 5.9 & 15.1 & \textbf{23.3} & \textbf{7.8} & \textbf{19.4} & 22.6 & 7.1 & 18.7 \\
hye & 35.4 & 23.3 & 32.0 & 18.8 & 7.1 & 14.7 & 11.4 & 4.8 & 8.5 & 16.3 & 6.2 & 14.3 & \textbf{20.5} & \textbf{8.5} & \textbf{17.5} \\
bos & 49.3 & 38.7 & 47.2 & 14.1 & 5.0 & 11.5 & 14.8 & 5.1 & 12.1 & 14.3 & 5.6 & 12.7 & \textbf{15.0} & \textbf{6.3} & \textbf{13.2} \\
kat & 50.0 & 40.3 & 48.7 & 11.4 & 5.3 & 10.2 & 10.9 & 4.9 & 9.9 & 9.7 & 4.3 & 9.3 & \textbf{13.2} & \textbf{7.2} & \textbf{12.6} \\
hat & 49.0 & 34.0 & 43.7 & 23.6 & 9.2 & 17.0 & 21.1 & 7.4 & 15.2 & 14.4 & 3.8 & 11.9 & \textbf{24.1} & \textbf{8.5} & \textbf{19.0} \\
khm & 44.3 & 39.8 & 44.4 & 5.5 & 1.8 & 5.2 & \textbf{8.3} & \textbf{4.8} & \textbf{8.0} & 3.4 & 1.1 & 3.3 & 3.7 & 1.2 & 3.6 \\
kur-k & 58.2 & 46.7 & 55.7 & 17.9 & 6.4 & 13.9 & 20.2 & 7.5 & 15.8 & 18.2 & 6.7 & 15.4 & \textbf{25.4} & \textbf{12.4} & \textbf{22.1} \\
kur-s & 35.9 & 23.3 & 35.3 & 13.3 & 5.5 & 12.3 & \textbf{21.9} & \textbf{13.2} & \textbf{19.4} & 14.7 & 4.6 & 13.3 & 16.6 & 5.4 & 15.1 \\
lao & 28.9 & 22.7 & 28.9 & 7.6 & 2.2 & 7.3 & 8.9 & 3.9 & 8.6 & \textbf{12.0} & \textbf{5.6} & \textbf{11.9} & 11.3 & 5.2 & 11.1 \\
mkd & 35.8 & 21.9 & 31.7 & 17.4 & 5.6 & 13.6 & 17.2 & 4.9 & 13.3 & 20.2 & 7.1 & 17.0 & \textbf{21.3} & \textbf{7.6} & \textbf{18.0} \\
nde & 49.6 & 40.3 & 48.5 & \textbf{18.4} & \textbf{9.9} & \textbf{16.3} & 17.1 & 9.1 & 15.4 & 14.2 & 8.1 & 13.3 & 14.1 & 8.0 & 13.5 \\
sna & 43.8 & 33.2 & 42.7 & 14.8 & 6.8 & 12.8 & 14.7 & 6.8 & 12.9 & 12.6 & 4.8 & 11.3 & \textbf{15.9} & \textbf{7.5} & \textbf{15.0} \\ \bottomrule
\end{tabular}
}
\caption{Comparison of different summarization approaches. Best scores in bold excluding oracle.}
\label{tab:comparison}
\end{table*}

Overall, we find that abstractive models fail to beat extractive ones for some languages, while extractive models and even the lead-3 baseline remain competitive for others.  
The fact that the baselines and extractive approaches still outperform abstractive neural models demonstrates the potential use of this corpus for further summarization research to improve abstractive models in less-resourced settings.

Results comparing the different approaches for 12 languages are shown in Table \ref{tab:comparison}.
The multilingual models tend to produce higher scores, likely due to positive transfer between languages.
However, the advantage is often only a few points beyond individual or extractive models.
The results of combining datasets (Table \ref{table:multilingual-lr-test-set}) show how LR-Sum can be combined with existing summarization datasets like XL-Sum to improve multilingual summarization model coverage. 
The additional data from the concatenation of LR-Sum and XL-Sum shows an expected advantage for languages not seen by the XL-Sum-only multilingual model.

\begin{table}[tb!]
\begin{center}
\resizebox{\linewidth}{!}{
\begin{tabular}{@{}lrcrrr@{}}
\toprule
 & \multicolumn{1}{c}{Training } &  & \multicolumn{1}{l}{} & \multicolumn{1}{l}{} & \multicolumn{1}{l}{} \\
Language & \multicolumn{1}{c}{Size} & In mT5 & \multicolumn{1}{c}{R1} & \multicolumn{1}{c}{R2} & \multicolumn{1}{c}{RL} \\ \midrule
sqi & 18,312 & \checkmark & 23.32 & 7.76 & 19.44 \\
hye & 920 & \checkmark & 16.27 & 6.18 & 14.31 \\
bos & 11,648 &  & 14.33 & 5.63 & 12.72 \\
kat & 511 & \checkmark & 9.71 & 4.28 & 9.26 \\
hat & 452 & \checkmark & 14.43 & 3.82 & 11.92 \\
khm & 3,888 & \checkmark  & 3.37 & 1.11 & 3.29 \\
kur-k & 791 & \checkmark & 18.24 & 6.73 & 15.38 \\
kur-s & 1,230 & \checkmark & 14.72 & 4.55 & 13.25 \\
lao & 11,964 & \checkmark & 12.00 & 5.62 & 11.85 \\
mkd & 1,223 & \checkmark & 20.20 & 7.14 & 17.03 \\
nde & 1,739 &  & 14.15 & 8.14 & 13.32 \\
sna & 383 & \checkmark & 12.63 & 4.81 & 11.35 \\ \bottomrule
\end{tabular}
}
\caption{Results of abstractive models for less-resourced languages of LR-Sum not also in XL-Sum from fine-tuning mT5 on LR-Sum data. Whether the languages are present in mT5 pre-training is marked with a check.}
\label{table:individual-results}
\end{center}
\end{table}

\subsection{Individual Model Results}
The results of training the individual models are shown in Tables \ref{tab:comparison} and \ref{table:individual-results}. 
The scores are generally slightly lower than the multilingual model with the exception of Albanian, Lao, and Northern Ndebele. 
The difference in training set size does not appear to be a factor in the performance, potentially because all the training set sizes for these less-resourced languages are small compared to the usual hundreds of thousands of examples found in datasets like MLSUM \citep{scialom-etal-2020-mlsum}. 
A language's presence in mT5's pre-training also does not appear to be indicative of better performance. 

\begin{table*}[tb!]
\centering
\resizebox{\linewidth}{!}{
\begin{tabular}{@{}lrrrrrrrrrrrrrrr@{}}
\toprule
 & \multicolumn{3}{c}{Oracle} & \multicolumn{3}{c}{Lead 3} & \multicolumn{3}{c}{LexRank} & \multicolumn{3}{c}{Luhn} & \multicolumn{3}{c}{TextRank} \\ \cmidrule(lr){2-4} \cmidrule(lr){5-7} \cmidrule(lr){8-10} \cmidrule(lr){11-13} \cmidrule(l){14-16}
Lang. & R1 & R2 & RL & R1 & R2 & RL & R1 & R2 & RL & R1 & R2 & RL & R1 & R2 & RL \\ \cmidrule(r){1-1} \cmidrule(lr){2-4} \cmidrule(lr){5-7} \cmidrule(lr){8-10} \cmidrule(lr){11-13} \cmidrule(l){14-16}
sqi & 43.90 & 29.71 & 39.58 & 19.45 & \textbf{6.75} & \textbf{14.96} & \textbf{19.57} & 5.85 & 3.52 & 18.25 & 6.13 & 3.88 & 16.66 & 5.13 & 3.06 \\
hye & 35.31 & 23.34 & 32.04 & \textbf{18.78} & \textbf{7.09} & \textbf{14.72} & 11.39 & 4.80 & 2.72 & 10.33 & 4.14 & 2.34 & 10.06 & 4.07 & 2.27 \\
bos & 49.27 & 38.72 & 47.19 & 14.13 & 4.98 & \textbf{11.48} & \textbf{14.78} & 5.09 & 3.32 & 13.81 & 5.21 & 3.56 & 13.30 & \textbf{5.30} & 3.74 \\
kat & 49.87 & 40.43 & 48.81 & \textbf{11.41} & \textbf{5.29} & \textbf{10.21} & 10.90 & 4.91 & 3.06 & 9.50 & 4.10 & 2.63 & 8.83 & 3.96 & 2.68 \\
hat & 48.92 & 34.06 & 43.76 & \textbf{23.63} & \textbf{9.22} & \textbf{16.98} & 21.13 & 7.43 & 4.24 & 19.75 & 7.67 & 4.65 & 18.52 & 6.90 & 4.12 \\
khm & 44.32 & 39.71 & 44.31 & 5.48 & 1.83 & \textbf{5.23} & \textbf{8.31} & \textbf{4.76} & 3.25 & 6.99 & 3.83 & 2.68 & 6.56 & 3.65 & 2.46 \\
kur-k & 58.31 & 46.66 & 55.46 & 17.94 & 6.37 & \textbf{13.89} & \textbf{20.24} & \textbf{7.55} & 4.84 & 18.42 & 7.23 & 4.69 & 16.82 & 6.57 & 4.38 \\
kur-s & 35.87 & 23.21 & 35.33 & 13.28 & 5.51 & \textbf{12.26} & \textbf{21.88} & \textbf{13.22} & 11.62 & 20.94 & 13.01 & 11.51 & 18.91 & 10.91 & 9.39 \\
lao & 29.00 & 22.66 & 29.01 & 7.60 & 2.18 & \textbf{7.31} & \textbf{8.92} & \textbf{3.90} & 2.03 & 7.73 & 3.36 & 1.78 & 7.36 & 3.20 & 1.74 \\
mkd & 35.79 & 21.86 & 31.76 & \textbf{17.44} & \textbf{5.64} & \textbf{13.63} & 17.24 & 4.90 & 2.65 & 14.86 & 3.73 & 1.74 & 14.27 & 3.54 & 1.66 \\
nde & 49.60 & 40.37 & 48.50 & \textbf{18.37} & \textbf{9.90} & \textbf{16.30} & 17.13 & 9.14 & 6.28 & 14.00 & 7.71 & 5.79 & 13.14 & 7.06 & 5.31 \\
sna & 43.83 & 33.13 & 42.59 & \textbf{14.78} & 6.78 & \textbf{12.77} & 14.73 & \textbf{6.80} & 4.14 & 12.49 & 5.62 & 3.67 & 11.19 & 4.51 & 2.58 \\ \bottomrule
\end{tabular}
}

\caption{Extractive model results on less-resourced languages that are not covered in XL-Sum. Results in bold are best or R1, R2, and RL across approaches excluding oracle.}
\label{table:extractive-results}
\end{table*}

\subsection{Extractive Results}
The results for extractive models can be found in Table \ref{table:extractive-results}. 
\texttt{Oracle} gives a sense of the upper bound that can be achieved through extractive models. 
The scores for the oracle are higher than both individual and multilingual abstractive models, which suggests there is plenty of room for improving performance for the abstractive baselines. 

For all the languages we evaluated, LexRank had higher scores than Luhn in terms of ROUGE-1, though Luhn was slightly higher in ROUGE-2 and ROUGE-L for Haitian Creole, Bosnian and Albanian. 
Lead-3 proves to be a strong baseline and scores higher than the  extractive models for RL and frequently for R1 and R2. 
In terms of R1, LexRank outperforms the individual abstractive models for Khmer, Georgian, Bosnian, Northern Ndebele, and Shona but ROUGE-L scores tend to be higher for the individual abstractive models.
The multilingual model still beats the lead-3 baseline except for Northern Ndebele and Khmer as shown in Table \ref{tab:comparison}.

\begin{table*}[tb!]
\centering
\begin{tabular}{@{}lcrrrrrr@{}}
\toprule
 &  & \multicolumn{3}{c}{LR- \& XL-Sum} & \multicolumn{3}{c}{XL-Sum} \\ \cmidrule(lr){3-5} \cmidrule(lr){6-8} 
Language & Not in XL-Sum & R1 & R2 & RL & R1 & R2 & RL \\ \midrule
Albanian & \checkmark & 22.55 & 7.07 & 18.72 & 8.98 & 0.94 & 7.69 \\
Amharic &  & 13.04 & 5.82 & 11.71 & 12.35 & 4.44 & 10.56 \\
Armenian & \checkmark & 20.49 & 8.47 & 17.46 & 0.41 & 0.15 & 0.41 \\
Azerbaijani &  & 15.99 & 8.33 & 15.19 & 14.44 & 5.89 & 13.34 \\
Bangla &  & 12.99 & 5.58 & 11.72 & 11.15 & 3.95 & 9.82 \\
Bosnian & \checkmark & 15.00 & 6.31 & 13.23 & 11.49 & 2.36 & 9.71 \\
Burmese &  & 28.69 & 14.51 & 26.13 & 2.07 & 0.43 & 1.99 \\
Dari Persian & \checkmark & 14.62 & 1.84 & 10.88 & 31.74 & 11.62 & 25.87 \\
Georgian & \checkmark & 13.20 & 7.17 & 12.60 & 0.09 & 0.00 & 0.09 \\
Haitian Creole & \checkmark & 24.09 & 8.46 & 18.98 & 13.23 & 3.19 & 10.97 \\
Hausa &  & 27.13 & 10.05 & 21.91 & 28.89 & 10.64 & 22.70 \\
Indonesian &  & 26.93 & 13.87 & 23.84 & 27.00 & 11.84 & 23.24 \\
Khmer & \checkmark & 3.67 & 1.17 & 3.62 & 0.42 & 0.15 & 0.40 \\
Kinyarwanda & \checkmark & 15.48 & 5.94 & 13.22 & 15.60 & 6.14 & 13.10 \\
Korean &  & 21.68 & 9.20 & 19.17 & 16.48 & 6.27 & 14.75 \\
Kurmanji Kurdish & \checkmark & 25.41 & 12.44 & 22.13 & 8.29 & 1.62 & 7.47 \\
Lao & \checkmark & 11.26 & 5.24 & 11.09 & 2.28 & 0.49 & 2.26 \\
Macedonian & \checkmark & 21.29 & 7.63 & 18.03 & 11.08 & 1.50 & 9.20 \\
Northern Ndebele & \checkmark & 14.14 & 8.05 & 13.55 & 3.91 & 1.20 & 3.76 \\
Pashto &  & 35.95 & 14.37 & 29.16 & 36.14 & 13.86 & 29.30 \\
Persian Farsi &  & 11.79 & 0.71 & 8.64 & 21.24 & 6.37 & 16.79 \\
Portuguese &  & 20.55 & 9.19 & 18.02 & 18.16 & 4.59 & 14.71 \\
Russian &  & 12.32 & 5.51 & 11.48 & 12.86 & 4.32 & 11.66 \\
Serbian &  & 16.63 & 5.07 & 14.03 & 15.24 & 3.32 & 12.42 \\
Shona & \checkmark & 15.88 & 7.50 & 14.99 & 4.79 & 1.27 & 4.49 \\
Somali &  & 28.80 & 11.54 & 24.30 & 31.39 & 13.15 & 26.21 \\
Sorani Kurdish & \checkmark & 16.60 & 5.44 & 15.08 & 5.76 & 0.46 & 5.14 \\
Swahili &  & 26.54 & 9.98 & 21.27 & 27.11 & 9.46 & 21.03 \\
Thai &  & 4.52 & 1.87 & 4.46 & 3.65 & 1.38 & 3.62 \\
Tigrinya &  & 13.07 & 3.70 & 11.30 & 12.79 & 3.50 & 10.80 \\
Turkish &  & 28.42 & 17.24 & 26.02 & 22.37 & 10.77 & 19.90 \\
Ukrainian &  & 14.83 & 6.84 & 13.28 & 14.71 & 5.42 & 13.05 \\
Urdu &  & 29.64 & 13.77 & 24.01 & 26.90 & 8.89 & 20.56 \\
Uzbek &  & 15.96 & 8.32 & 14.51 & 12.61 & 4.13 & 11.36 \\
Vietnamese &  & 25.06 & 14.13 & 21.52 & 26.51 & 13.67 & 21.62 \\ \bottomrule
\end{tabular}
\caption{Results from a multilingual model trained on both LR-Sum and XL-Sum data compared with a multilingual model trained only on XL-Sum. We additionally omit Tibetan and Greek from the results as they have only enough data for test sets.
Higher-resourced languages are also omitted.}
\label{table:multilingual-lr-test-set}
\end{table*}

\subsection{Multilingual Model Results}

Table \ref{table:multilingual-lr-test-set} shows the results of mT5 \citep{xue-etal-2021-mt5} trained on the concatenation of training data from LR-Sum and XL-Sum compared with the model checkpoint of mT5 trained on XL-Sum only. 
As expected, languages not present in XL-Sum had much better performance with the model trained on both datasets. 
Dari Persian did not perform better likely due to Farsi already being represented in XL-Sum and the two languages being very similar.
Scores for Greek and Tibetan were effectively zero as there is only enough data in LR-Sum for a test set and so there was no training data in those languages due to data scarcity. 

The results for additional training data for languages present in both languages are more mixed. 
Despite both datasets being news data, it is possible there are differences in dialect, topic, or standardization that account for the differences.
We discuss the performance of the two multilingual models evaluated on the XL-Sum test set in Appendix \ref{sec:multiling-xl-sum-eval}.

\section{Conclusions and Future Work}
We have presented LR-Sum, a permissively-licensed summarization dataset for less-resourced languages based on news data. 
We have demonstrated LR-Sum's usefulness in augmenting the training data of other multilingual summarization models and demonstrated potential for further research in summarization for less-resourced languages.
Even with the best performing model, the results are only slightly higher than the lead-3 baseline, which indicates ample room for improvement and future research directions. 

In future work, we plan to experiment with leveraging additional training data like the remaining portions of the MOT data which were not suitable for extracting summaries but may still be useful in fine-tuning a multilingual language model to perform better on certain less-resourced languages.
LR-Sum also presents opportunities for few- and zero-shot experimentation for languages where there are not enough examples to use as training data, but where the data that does exist may be useful as a test set.
We look forward to collaborating with speakers of the languages included in LR-Sum to further increase the quality and quantity of summarization data for less-resourced languages. 

\section*{Limitations}
A limitation of this work is that the dataset has not yet been thoroughly vetted by native speakers of the languages contained in the dataset.
We acknowledge the importance of working with native speakers and manually reviewing datasets in greater detail as argued for by \citet{kreutzer-etal-2022-quality} and \citet{lignos-etal-2022-toward}. 
We hope to do more manual review of LR-Sum and other summarization datasets in the near future.

\section*{Ethics Statement}
Our work provides a dataset for further research on summarization for less-resourced languages. 
Automatic summarization has the potential to assist users in digesting information. 
It is our intention that providing a summarization dataset with coverage of less-resourced languages will benefit speakers of languages that may otherwise not have had access to this technology. 

However, there is also cause for caution.
The results of our work used automatic evaluation metrics and generated summaries have not yet been subjected to more rigorous human review.
Even just based on automated metrics, it is clear there is still room for improvement of the models as they tend to score lower than higher resourced counterparts on similar tasks.
Therefore, the models presented in this work should be considered baselines for further work.
The dataset and models presented in this work are meant to support further research in summarization of less-resourced languages and not intended for immediate deployment in applications. 

In particular, the abstractive summarization models, like most text generation models, have the potential to make factual errors, which have the potential to mislead or misinform.
Additionally, both extractive and abstractive models may lack adequate context or miss important information.
As mentioned in the limitations section, this dataset, like most summarization news datasets, has not been fully manually reviewed and so may contain a few erroneous summaries despite our  best efforts.

\section*{Acknowledgments}
Chester Palen-Michel was supported by a grant from eBay while performing this work.

\FloatBarrier

\bibliography{anthology,custom}
\bibliographystyle{acl_natbib}

\clearpage

\appendix
\section{Hyper-parameter Settings}
\label{sec:hyperparams}
The hyper-parameters for both individual models and multilingual models were chosen following \citet{hasan-etal-2021-xl}. 

\subsection{Individual Models}
\label{sec:hyperparams-mono}
For hyperparameters in training each individual model we use 
a learning rate of 5.0e-4, per device train batch size of 2, 16 gradient accumulation steps, 100 warm-up steps, a maximum input length of 512, a maximum inference length of 84, a beam size of 4, no repeat ngram size of 2, length penalty of 0.6, label smoothing factor of 0.1 and weight decay of 0.01. 
We train for 10 epochs. 
Training time was roughly 2 days in total to train the 12 individual models.

\subsection{Multilingual Model}
\label{sec:hyperparams-multi}
We use a learning rate of 1.0, 5000 warmup steps, weight decay of 0.01, per device train batch size of 2, 16 gradient accumulation steps, maximum steps of  50,000, a label smoothing factor of 0.1, and an upsampling factor of 0.5. 
We trained on two NVIDIA GeForce RTX 3090 GPUs. 
Training time was roughly 3 days.

\section{Evaluating Multilingual Models on XL-Sum}
\label{sec:multiling-xl-sum-eval}
We additionally evaluated the two multilingual models on the XL-Sum test data.
The results can be seen in Table \ref{table:multilingual-xl-test-set}. 
We found that the addition of LR-Sum data did not have a positive impact on performance but instead tended to degrade model performance slightly. 
We speculate that despite both being news summarization datasets there could be some amount of difference in content or style that accounts for the slightly lower performance. 
Another plausible explanation could be that adding relatively small amounts of data for additional languages degrades performance due to the model's limited capacity to add additional languages.

\begin{table*}[tb!]
\centering
\begin{tabular}{@{}lcrrrrrr@{}}
\toprule
 &  & \multicolumn{3}{c}{LR- \& XL-Sum} & \multicolumn{3}{c}{XL-Sum} \\ \cmidrule(lr){3-5} \cmidrule(lr){6-8}  
Language & In LR-Sum & R1 & R2 & RL & R1 & R2 & RL \\ \midrule
Amharic & \checkmark & 18.21 & 6.87 & 16.45 & 20.08 & 7.41 & 18.06 \\
Azerbaijani & \checkmark & 18.73 & 8.00 & 17.15 & 21.37 & 9.54 & 19.35 \\
Bengali & \checkmark & 21.79 & 8.95 & 19.03 & 24.35 & 10.12 & 21.25 \\
Burmese &  & 15.16 & 4.49 & 13.75 & 16.17 & 5.15 & 14.41 \\
Gujarati &  & 20.55 & 7.15 & 18.59 & 21.94 & 7.74 & 19.91 \\
Hausa & \checkmark & 37.27 & 16.07 & 29.85 & 39.42 & 17.67 & 31.64 \\
Hindi &  & 34.88 & 14.45 & 28.93 & 36.91 & 16.32 & 30.88 \\
Igbo &  & 27.46 & 8.30 & 21.13 & 31.66 & 10.21 & 24.56 \\
Indonesian & \checkmark & 35.03 & 15.70 & 29.13 & 36.99 & 17.02 & 30.74 \\
Japanese &  & 39.10 & 23.17 & 31.61 & 41.71 & 25.19 & 33.65 \\
Kirundi &  & 29.77 & 12.65 & 23.80 & 31.99 & 14.44 & 25.82 \\
Korean & \checkmark & 21.89 & 10.51 & 20.54 & 23.76 & 11.53 & 22.42 \\
Kyrgyz &  & 16.24 & 7.17 & 14.72 & 18.36 & 8.02 & 16.46 \\
Marathi &  & 20.48 & 8.64 & 18.51 & 22.05 & 9.54 & 20.02 \\
Nepali &  & 24.55 & 9.12 & 22.25 & 26.58 & 10.22 & 24.24 \\
Oromo &  & 16.37 & 5.42 & 14.40 & 18.75 & 6.22 & 16.16 \\
Pashto & \checkmark & 36.30 & 13.74 & 29.71 & 38.25 & 15.48 & 31.74 \\
Persian & \checkmark & 33.47 & 13.22 & 26.86 & 35.71 & 15.06 & 29.12 \\
Portuguese & \checkmark & 33.52 & 13.85 & 25.91 & 35.29 & 15.39 & 27.50 \\
Punjabi &  & 28.82 & 10.73 & 23.92 & 30.80 & 12.18 & 25.56 \\
Russian & \checkmark & 22.75 & 9.10 & 19.31 & 25.28 & 10.78 & 21.51 \\
Scottish Gaelic &  & 27.37 & 9.81 & 22.00 & 29.04 & 10.95 & 22.89 \\
Serbian-cyrillic &  & 21.07 & 6.48 & 17.84 & 23.76 & 7.98 & 20.15 \\
Serbian-latin & \checkmark & 20.58 & 5.70 & 17.14 & 21.64 & 6.68 & 18.24 \\
Sinhala &  & 20.73 & 7.95 & 17.99 & 21.47 & 8.06 & 18.85 \\
Somali & \checkmark & 30.13 & 10.54 & 22.97 & 31.52 & 11.53 & 24.21 \\
Swahili & \checkmark & 36.52 & 17.14 & 29.72 & 37.67 & 17.86 & 30.94 \\
Tamil &  & 22.54 & 9.97 & 20.56 & 24.33 & 11.03 & 22.06 \\
Telugu &  & 16.12 & 5.26 & 14.44 & 17.72 & 5.72 & 15.84 \\
Thai & \checkmark & 10.34 & 4.07 & 9.90 & 12.28 & 4.78 & 11.87 \\
Tigrinya & \checkmark & 23.01 & 7.05 & 19.21 & 25.25 & 7.99 & 21.08 \\
Turkish & \checkmark & 26.07 & 11.95 & 23.56 & 28.90 & 13.79 & 26.15 \\
Ukrainian & \checkmark & 22.06 & 8.90 & 19.24 & 23.99 & 10.14 & 20.92 \\
Urdu & \checkmark & 37.39 & 16.48 & 30.79 & 39.48 & 18.33 & 32.81 \\
Uzbek & \checkmark & 15.52 & 5.58 & 14.18 & 16.82 & 6.35 & 15.35 \\
Vietnamese & \checkmark & 28.11 & 12.80 & 22.04 & 30.26 & 14.38 & 24.14 \\
Welsh &  & 30.43 & 9.73 & 24.46 & 32.62 & 11.61 & 26.12 \\
West African Pidgin &  & 36.54 & 14.29 & 28.56 & 37.98 & 15.11 & 29.86 \\
Yoruba &  & 29.45 & 10.36 & 23.11 & 31.62 & 11.66 & 25.06 \\ \bottomrule
\end{tabular}
\caption{Results evaluating on the XL-Sum test set. 
Results from an mT5 multilingual model fine-tuned on both LR-Sum and XL-Sum data compared with an mT5 multilingual model fine-tuned only on XL-Sum. Scores are ROUGE-1 and 2 (R1, R2) and ROUGE-L (RL). 
Higher-resourced languages are also omitted.}
\label{table:multilingual-xl-test-set}
\end{table*}

\section{Dataset Splits}
Table \ref{tab:splits} shows the dataset splits as described in Section \ref{sec:splits}

\begin{table*}[tbh]
\centering
\begin{tabular}{llrrr}
\toprule
Language & ISO 639-3 & \multicolumn{1}{l}{Train} & \multicolumn{1}{l}{Validation} & \multicolumn{1}{l}{Test} \\
\midrule
Albanian & sqi & 18,312 & 2,289 & 2,289 \\
Amharic & amh & 0 & 0 & 154 \\
Armenian & hye & 920 & 500 & 500 \\
Azerbaijani & aze & 6,487 & 810 & 811 \\
Bangla & ben & 0 & 0 & 715 \\
Bosnian & bos & 11,648 & 1,455 & 1,456 \\
Burmese & mya & 7,921 & 990 & 990 \\
Chinese Simplified & cmn & 2,103 & 500 & 500 \\
Chinese Traditional & cmn & 483 & 500 & 500 \\
Dari Persian & prs & 12,037 & 1,504 & 1,505 \\
English & eng & 20,976 & 2,621 & 2,622 \\
French & fra & 1,126 & 500 & 500 \\
Georgian & kat & 511 & 500 & 500 \\
Greek & ell & 0 & 0 & 583 \\
Haitian Creole & hat & 452 & 500 & 500 \\
Hausa & hau & 0 & 0 & 390 \\
Indonesian & ind & 968 & 500 & 500 \\
Khmer & khm & 3,888 & 486 & 486 \\
Kinyarwanda & kin & 0 & 0 & 698 \\
Korean & kor & 10,499 & 1,312 & 1,312 \\
Kurmanji Kurdish & kur & 791 & 500 & 500 \\
Lao & lao & 11,964 & 1,495 & 1,496 \\
Macedonian & mkd & 1,223 & 500 & 500 \\
Northern Ndebele & nde & 1,739 & 500 & 500 \\
Pashto & pus & 16,854 & 2,106 & 2,107 \\
Persian Farsi & fas & 10,744 & 1,342 & 1,343 \\
Portuguese & por & 643 & 500 & 500 \\
Russian & rus & 10,812 & 1,351 & 1,351 \\
Serbian & srp & 4,974 & 621 & 622 \\
Shona & sna & 383 & 500 & 500 \\
Somali & som & 0 & 0 & 165 \\
Sorani Kurdish & kur & 1,230 & 500 & 500 \\
Spanish & spa & 2,544 & 500 & 500 \\
Swahili & swh & 0 & 0 & 588 \\
Thai & tha & 2,278 & 500 & 500 \\
Tibetan & bod & 0 & 0 & 182 \\
Tigrinya & tir & 0 & 0 & 115 \\
Turkish & tur & 28,672 & 3,583 & 3,584 \\
Ukrainian & ukr & 5,784 & 722 & 723 \\
Urdu & urd & 10,847 & 1,355 & 1,356 \\
Uzbek & uzb & 9,568 & 1,195 & 1,196 \\
Vietnamese & vie & 11,676 & 1,459 & 1,460 \\
\bottomrule
\end{tabular}
\caption{Train, validation, and test split sizes for LR-Sum by language.}
\label{tab:splits}
\end{table*}



\end{document}